\documentclass[3p,review,12pt]{elsarticle}

\usepackage{lineno,hyperref}
\usepackage[version=4]{mhchem}
\usepackage{graphicx}
\usepackage{amsmath}
\usepackage{array}
\usepackage{hyperref}
\usepackage{subcaption}
\modulolinenumbers[5]
\linenumbers               
\let\linenumbers\nolinenumbers\nolinenumbers

\journal{Applied Energy}

\biboptions{numbers, sort&compress}
\begin{document}

\begin{frontmatter}

\title{Machine Vision for Natural Gas Methane Emissions Detection Using an Infrared Camera}

%% Group authors per affiliation:
\author[ere]{Jingfan Wang\corref{cor}}
\ead{jingfan@stanford.edu}

\author[cs]{Lyne P. Tchapmi \fnref{fn1}}

\author[ere]{Arvind P. Ravikumar \fnref{fn1,fn2}}

\author[csu]{Mike McGuire \fnref{fn3}}

\author[csu]{Clay S. Bell}

\author[mecsu]{Daniel Zimmerle}

\author[cs]{Silvio Savarese}

\author[ere]{Adam R. Brandt}

\cortext[cor]{Corresponding author}
\fntext[fn1]{L.P.T. and A.P.R. contributed equally to this work.}
\fntext[fn2]{Current affiliation is Harrisburg University of Science and Technology.}
\fntext[fn3]{Current affiliation is SeekOps Inc.}

\address[ere]{Department of Energy Resources Engineering, Stanford University, 367 Panama St., Stanford, 94305, California, United States}
  
\address[cs]{Department of Computer Science, Stanford University, 353 Serra Mall, Stanford, 94305, California, United States}

\address[csu]{Colorado State University Energy Institute, 430 North College Av., Fort Collins, 80542, Colorado, United States}

\address[mecsu]{Department of Mechanical Engineering, Colorado State University, 1374 Campus Delivery, Fort Collins, 80523,
Colorado, United States}

\begin{abstract}
In a climate-constrained world, it is crucial to reduce natural gas methane emissions, which can potentially offset the climate benefits of replacing coal with gas. Optical gas imaging (OGI) is a widely-used method to detect methane leaks, but is labor-intensive and cannot provide leak detection results without operators' judgment. In this paper, we develop a computer vision approach to OGI-based leak detection using convolutional neural networks (CNN) trained on methane leak images to enable automatic detection. First, we collect $\sim$1 M frames of labeled video of methane leaks from different leaking equipment for training, validating and testing CNN model, covering a wide range of leak sizes (5.3-2051.6 g\ce{CH4}/h) and imaging distances (4.6-15.6 m). Second, we examine different background subtraction methods to extract the methane plume in the foreground. Third, we then test three CNN model variants, collectively called GasNet, to detect plumes in videos taken at other pieces of leaking equipment. We assess the ability of GasNet to perform leak detection by comparing it to a baseline method that uses optical-flow based change detection algorithm. We explore the sensitivity of results to the CNN structure, with a moderate-complexity variant performing best across distances. We find that the detection accuracy (fraction of leak and non-leak images correctly identified by the algorithm) can reach as high as 99\%, the overall detection accuracy can exceed 95\% for a case across all leak sizes and imaging distances. Binary detection accuracy exceeds 97\% for large leaks ($\sim$710 g\ce{CH4}/h) imaged closely ($\sim$5-7 m). At closer imaging distances ($\sim$5-10 m), CNN-based models have greater than 94\% accuracy
across all leak sizes. At farthest distances ($\sim$13-16 m), performance degrades rapidly, but it can achieve above 95\% accuracy to detect large leaks (\textgreater 950 g\ce{CH4}/h). The GasNet-based computer vision approach could be deployed in OGI surveys to allow automatic vigilance of methane leak detection with high detection accuracy in the real world.   
\end{abstract}

\begin{keyword}
natural gas\sep methane emission\sep deep learning\sep convolutional neural network\sep computer vision\sep optical gas imaging
\end{keyword}

\end{frontmatter}

\linenumbers

\section{Introduction}
Natural gas plays a significant role in the global energy system. In the United States, natural gas is now the largest single energy resource, accounting for 31.8\% of energy production in 2017 \cite{eia_2017}. Advanced and cost-effective drilling and production techniques like hydraulic fracturing and horizontal drilling have stimulated increased natural gas production from shale formations \cite{eia_2017, eia_2016}. The coupling between natural gas systems and electricity systems has been increasing with rising deployment of gas-fired power plants and gas-fired combined heat and power systems \cite{qiao2017interval, qadrdan2015impact, chaudry2014combined, chiang2016large, touretzky2016effect, mongibello2016comparison}. Increasing residential and commercial heating demand also contributes to growing natural gas consumption and production \cite{spoladore2016model}. This expansion has had climate benefits: due to decreases in the cost of gas and renewables, coal consumption for power generation in the U.S. has dropped from 39.0\% in 2010 to 29.9\% in 2017 \cite{eia_2018}. Furthermore, gas use in the power sector will likely continue to rise even in high renewable energy systems: spare capacity of fast-ramping natural gas power plants can compensate for variability introduced by wind and solar plants \cite{rothleder2017}.

However, there is debate over the role of natural gas in a low-carbon future \cite{howarth2011methane, wigley2011coal, alvarez2012greater, burnham2011life, jackson2014natural, zhang2016climate}. Upon combustion, natural gas causes lower climate and air quality damages than coal. However, the loss of natural gas may exert a negative effect on the climate due to the high global warming potential (GWP) of methane (36 times more potent per kg than \ce{CO2} over 100 years \cite{eia_2017understanding}). Recent experimental work has highlighted important features of methane emissions from the oil and gas industry. First, methane emissions are generally underestimated by official inventories \cite{brandt2014methane, alvarez2018assessment}. Second, experiments should be designed to understand large but intermittent or infrequent sources, which could make up a large portion of emissions, even though they are hard to sample during conventional surveys \cite{vaughn2018temporal, schwietzke2017improved}. Finally, natural gas emission volumes are highly skewed: the largest 5\% of leaks account for approximately 50\% of the total emissions \cite{brandt2016methane}.

Regulatory approaches to reduce fugitive emissions in the US and Canada typically require periodic leak detection and repair (LDAR) surveys at oil and gas facilities. LDAR surveys typically use EPA (Environmental Protection Agency) ‘Method-21’ or manually-operated infrared (IR) optical gas imaging (OGI) technologies to detect leaks \cite{eia_2018method, eia_2016new}.  An open-source model - the Fugitive Emissions Abatement Simulation Toolkit (FEAST) - allows simulation-based comparison of different LDAR technologies and programs \cite{kemp2016comparing}.

Because of its ease of use, IR OGI cameras have become the most commonly used LDAR technology. Gas plumes can be visualized in the IR camera: absorptive plumes look black and emissive plumes look white. Despite its widespread use, OGI performance is affected by environmental conditions, operator experience, survey practices, leak size distributions and gas composition \cite{ravikumar2016optical}. We recently experimentally derived detection probability curves for OGI-based methane leak detection under real-world conditions and found that the median and 90\% detection likelihood limit follow a power-law relationship with imaging distance \cite{ravikumar2018good}.

Despite the usefulness of OGI, a number of fundamental challenges exist: (1) labor costs for manual OGI surveys are high, (2) continuous monitoring with IR cameras is infeasible, (3) IR cameras cannot provide real-time feedback of leak detection results without operators' judgement, and (4) the quality of survey varies between different OGI operators.

To address these challenges, in this paper we explore computer vision approaches based on convolutional neural networks (CNNs) that are trained to examine IR images to determine whether there is a methane leak or not. Such computer vision approaches, if successful, would allow automatic leak detection and remove uncertainty associated with operator experience. There are few automated leak detection products currently available, and there is a lack of scientific and systematic analysis of the limits and effectiveness of automated OGI-based technology. We aim to fill some of these gaps in this paper.

Our method for automatic detection from IR images of methane leaks was developed in five stages.  First, we build a video dataset of methane leaks, GasVid, which includes large numbers of labeled videos of methane leaks with associated leak sizes from different leak locations and imaging distances. Second, we systematically test background subtraction methods and develop a CNN model called GasNet to identify leaks from video frames.  Third, we derive the probability of correct assessment by the automatic OGI-based methane detection under different imaging distances, leak sizes and environmental conditions. Fourth, we test three CNN model variants and compare the results from the CNN-based analysis with conventional optical flow algorithms. Finally, we also compute the detection accuracy (fraction of leak and non-leak images correctly identified by the algorithm) results for the dataset at each distance and for the entire dataset among all distances.

\section{Related work}
Currently, very few commercial products allow for automatic leak detection using video imagery. The leading IR camera vendor, FLIR systems, in partnership with Providence Photonics, is actively developing an add-on to the existing FLIR IR camera that quantifies leak rates \cite{flir}. This product makes use of the temporal motion of the plume and the apparent temperature difference between the hydrocarbon plume and the background to detect the leak. It has a multi-stage algorithm to confirm an emission event \cite{abdel2013remote}. Another company, Rebellion Photonics, uses a more sophisticated hyper-spectral imaging camera that can distinguish between different gas species and claims to quantify the leak size in real time \cite{rebellion}.

There are a large number of studies on smoke detection published by researchers in the field of image processing and computer vision. Methane plumes share many similarities with smoke in terms of non-rigidity, dispersion, color attenuation and blending, and irregularities in motion. But methane plumes and smoke are captured by different types of cameras: methane plumes are seen in grayscale and can only be seen by infrared cameras; smoke is imaged in RGB (red, green and blue) colors and can be seen by conventional visual spectrum cameras. 

Common image-based smoke emission detection approaches include color modeling, change detection, texture analysis and machine learning models \cite{hsu2018industrial}. The four models are reviewed and compared in the following discussion.

The color modeling method is based on color saturation or image intensity value distribution. Smoke pixels can be detected as they have relatively lower color saturation \cite{ccelik2007fire, lee2012smoke}. 

Change detection algorithms are used to automatically detect changes or movements \cite{radke2005image}. Regions containing moving features such as leaking gas are separated from regions that are more static, i.e. the background. With a sequence of images, backgrounds are assessed and subtracted from the original image, and a threshold is applied to the residual image to produce a binary image \cite{cheung2005robust,collins2000system}. Assessment of the background can be supported by probabilistic distribution of the background pixels assuming hybrid Gaussian distributions \cite{friedman1997image, stauffer1999adaptive}. Examining the optical flow field entropy and glimmering pixels outlining the edge can also help determine the background \cite{kopilovic2000application, toreyin2005wavelet}.

Texture analysis can utilize either a single image or a sequence of images. Feature vectors are gained by performing wavelet transform or other texture descriptors, providing inputs for classifier training \cite{calderara2008smoke, gubbi2009smoke}. 

Lastly, machine vision techniques for smoke detection are also discussed in the literature. Application of CNNs in detecting smoke from wildfires is discussed in the paper by \cite{hohberg2015wildfire}.

Deep learning has been widely used in the energy and environmental field, including solar prediction \cite{sun2018solar, gensler2016deep}, wind energy forecasting \cite{wang2017deep, yu2019scene}, electricity price forecasting \cite{lago2018forecasting}, building energy forecasting and optimization \cite{fan2017short, guo2018deep, rahman2018predicting}, etc. However, there is a paucity of research on application of deep learning in methane emission detection.

In this paper, we will use deep learning approaches to perform image analysis of methane leaks imaged using a FLIR GF-320 infrared camera. Instead of using hand-crafted features in traditional machine learning approaches, deep learning automatically calculates hierarchical features of the input data in order to accomplish tasks such as object detection, speech recognition, video classification, etc. Usually in deep learning, there are multiple layers. The first several layers learn low-level features, such as points, edges and curves in the input data. Deeper layers can learn higher-level features that help the algorithm perform in a way similar to a human thought process \cite{zhang2003low}. CNN, a type of deep architecture, have been found to be highly powerful in image recognition and object detection \cite{szegedy2015going}. There are many sophisticated CNN models, such as AlexNet, VGG, Inception, and ResNet \cite{szegedy2015going, krizhevsky2012imagenet, simonyan2014very, he2016deep}. While these models differ in many respects, they all consist of an input layer, several hidden layers and an output layer. The hidden layers usually have Conv-Pool ( layer-pooling layer) structures, in which layers apply a convolution operation to the input and deliver the result to the next layer, and then pooling layers subsample the output of the layer and reduce the dimensionality of the data. The hidden layer could also be a fully-connected layer to connect every neuron in the former layer to every neuron in the latter layer. At the end, the output layer, usually a fully-connected layer, estimates the regression result or the classification score for each class.

\section{Datasets-GasVid}
CNNs require large numbers of data samples to train a network of deep learning. Therefore, in order to apply CNN to detecting methane leaks, we have begun building a large dataset named GasVid, which includes labeled videos of methane leaks from various leakage sources covering wide range of leak sizes.  In all GasVid video segments, the actual leak rate is known and listed. Videos were taken across a range of environmental conditions, camera orientations, and imaging distances, representing a realistic range of leak scenarios. 
We generated GasVid with controlled-release experiments at the Methane Emissions Technology Evaluation Center (METEC) at Colorado State University in Fort Collins from July 10 - July 14 2017. METEC, funded through the ARPA-E\textsc{\char13}s Methane Observation Networks with Innovative Technology to Obtain Reduction (MONITOR) program, is a controlled-release test facility that mimics real-world gas leaks found at natural gas production sites. At the time of the study, METEC contained \textgreater50 metered leak sources made from $1/4''$ steel tubing. Operators control the flow rate by adjusting the orifices and the regulated gas pressure controlled at a central pressure regulator connected to high pressure compressed natural gas.
All videos were taken at METEC by the authors J.W. and A.P.R. Between July 10 and July 14, we collected a total of 31 24-min videos at separators (frame rate is $\sim$15 frames per second, resulting in $\sim$669,600 frames in total were recorded). We collected videos not only from separators, but also from tanks. At tank leaks, a total of $\sim$345,600 frames were recorded, but poor orientation of the leak and camera location rendered these videos un-usable. Separator videos were recorded at two point-source leak locations: separator on pad 1 (13 videos), separator on pad 2 (18 videos) at imaging distances between 4.6 m and 15.6 m. Data from separator 2 (separator on pad 2) is used for training and validating the CNN models, while data from separator 1 (separator on pad 1) is exclusively used for testing. In all experiments, a tripod-mounted FLIR GF-320 IR camera operating in the normal mode (not high sensitivity mode) was used to record the leak. Each video was taken with a unique combination of distance and camera orientation.  There are 5 imaging distances in total: 4.6 m, 6.9 m, 9.8 m, 12.6 m and 15.6 m. Within the 24-min-video, eight size classes (including zero) of leaks with flow rate ranging from 5.3 - 2051.6 g\ce{CH4}/h (0.3 - 124.4 standard cubic feet per hour, or scfh) were recorded in 3-min interval as shown in Table 1. 
\begin{table*}
    \centering
    \caption{Leak rates and the associated leak classes recorded from each imaging distance and leak source. Leak rates reported with error at 95\% confidence interval (CI).}
    \label{class_rate}
    \begin{tabular}{c|c|c}
    \hline
      \textbf{Leak class label} & \textbf{Leak rate in scfh ($\pm$95\% CI)} & \textbf{Leak rate in g/h ($\pm$95\% CI)} \\
      \hline
      Class 0 & 0.3 $\pm$ 0.0 & 5.3 $\pm$ 0.1 \\
      Class 1 & 16.8 $\pm$ 0.1 & 277.7 $\pm$ 1.1 \\
      Class 2 & 43.2 $\pm$ 0.2 & 713.1 $\pm$ 2.6 \\
      Class 3 & 58.1 $\pm$ 0.2 & 958.8 $\pm$ 3.1 \\
      Class 4 & 68.1 $\pm$ 0.3 & 1124.3 $\pm$ 4.3 \\
      Class 5 & 84.2 $\pm$ 0.3 & 1389.8 $\pm$ 4.8 \\
      Class 6 & 109.5 $\pm$ 2.5 & 1806.1 $\pm$ 41.4 \\
      Class 7 & 124.3 $\pm$ 2.9 & 2051.6 $\pm$ 48.0 \\
      \hline
    \end{tabular}
\end{table*}

In every 3-min leak video, the plume may not be steady at the beginning and the end. Therefore, we cut the first 15 seconds and the last 5 seconds of each 3-min-video. One frame of the video has a dimension of 240x320x1, indicating that the image is in grayscale.

As this is a first study, we include simple testing conditions. A tripod was used to avoid extraneous camera movement which may confuse the algorithm. We also exclude interference from cars, people. Our videos do not include moving vegetation, vapor and steam, which are possible in the real world. In addition, the largest leak does not occupy the entire field of view of the camera even at the shortest imaging distance, allowing us to capture the shape of the entire plume. Future studies can be performed to relax these idealities. 

\begin{figure}[!ht]
    \centering
    \includegraphics{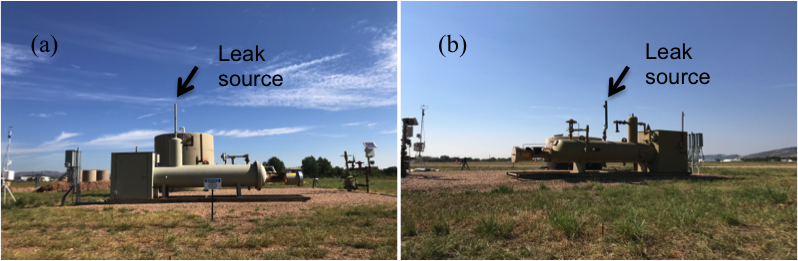}
    \caption{(a) Image of a separator on pad 1 with an imaging distance of 9.8 m. (b) Image of a separator on pad 2 with an imaging distance of 9.8 m. The photos were taken by iPhone which was directly next to the infrared camera shot.}
    \label{separators}
\end{figure}

\begin{figure}[!ht]
    \centering
    \includegraphics[width = 15cm]{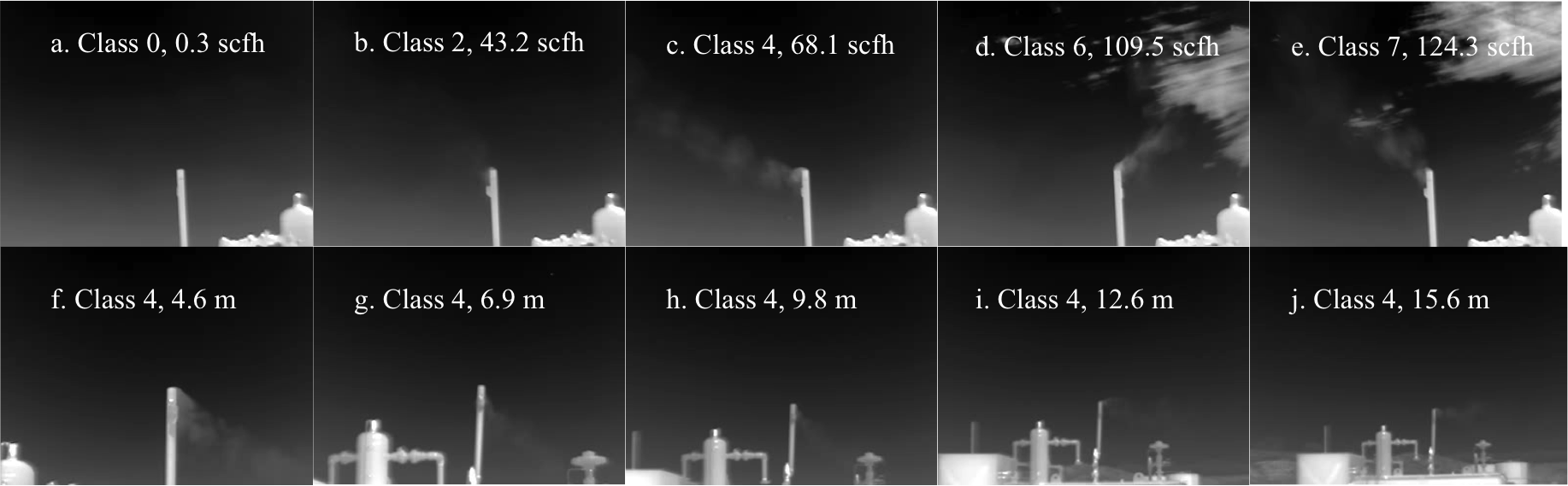}
    \caption{(a-e): Representative frames showing leak scenarios with five different leak sizes. Note that clouds have moved into the field of view of the camera during measurement of class-6 and class-7 leaks. (f-j): Representative frames of leak scenarios with leak class 4 (68.1 scfh) from five different imaging distances.}
    \label{leaksize}
\end{figure}

Figure 1a and 1b illustrate the leak locations on separator 1 and separator 2 respectively. The top row of images in Figure 2 shows representative frames from five different leak classes on separator 2, with associated leak rates. The bottom row of images in Figure 2 shows representative frames from a class-4 leak on separator 2 at different imaging distances. As the imaging distance increases, the plume becomes challenging to observe.

All the videos were recorded directly on the camera itself instead of using FLIR ResearchIR Software (an IR camera control and analysis software). This lowers the resolution of the videos due to on-camera compression. There were also occasional clouds appearing in the field of view of the camera (see Figure 2d and 2e). Usually clouds are moving and changing at a low speed and thus clouds would ideally be removed by a suitable background subtraction method. Lastly, wind speed and wind orientation affect the appearance of methane plume on images and were not controlled during the experiment. 

\section{Method}
In this section, we outline the workflow of our automatic detection algorithm. We introduce three background subtraction methods, one image normalization method, and three CNN model variants. We also explicitly explain the CNN model. Finally, a baseline model that does not use CNN is described as a point of comparison for the accuracy of the results.

\subsection{Workflow}
The probability curve of correct assessment by the automated OGI-based technology (a systematic way to examine the technology efficacy) is generated with 7 possible binary classification cases. Each binary classification classifies an image as a non-leak image or a leak image from one of the 7 leak classes. This is performed at each of the 5 imaging distances. Thus, in our base case, there are 35 cases representing the binary classification accuracy results of 35 independent trained experiments under different distances and leak sizes on the curve (we explore training with all datasets across leak sizes and distances at once below).

Each of 35 independent trained experiments is conducted using the same workflow. Still frames are first extracted from the videos and coupled with the associated class labels. The extracted images are pre-processed through background subtraction and image normalization. The processed images are individually fed into the trained CNN detection module. Finally, the accuracy of prediction is calculated as the fraction of images correctly identified by the algorithm.

\subsection{Background subtraction methods}

Background subtraction is the process of extracting the image foreground, which is the plume in our case, for further analysis. An idealized background subtraction method would create an image containing only the plume and no other objects. This would allow the algorithm to localize the plume easily and make the training process much faster and more accurate. All real background subtraction methods result in some non-plume features remaining in the image.

We systematically test three background subtraction methods and compare the results with a baseline method without any background subtraction. Three background subtraction methods are: (1) fixed background subtraction; (2) moving average background subtraction; and (3) Mixture of Gaussians-based (MOG) background subtraction.

In the GasVid dataset, each of the 31 24-min videos contains a 3-min non-leak video (class 0) at the beginning. For the fixed background subtraction method, we use the average of all the frames from the class-0 segment (i.e., non-leaking segment) as the background image for the corresponding 24-min video at a given imaging distance and camera orientation. Thus, in the fixed background case, every leak frame from the video is assumed to have the same background scene.

Instead of having a fixed background for all frames in one video, we can generate a moving average background for every frame in the video \cite{hsu2018industrial}. Our method creates a background image for each frame as the median of the previous 210 images. This is equivalent to the median frame from a moving lagged 14-second-long video. The idea behind moving average background is that smoothing out plume variations over a multi-second period, we can subtract the background to emphasize the frame-specific variation in the plume.

The Mixture of Gaussians (MOG)-based background subtraction involves learning a probabilistic model of each pixel using an appropriate number of Gaussian distributions of pixel intensities to identify static and moving pixels or colors \cite{opencv}. We use an adaptive background mixture model which chooses the appropriate number of Gaussian distribution for each pixel \cite{zivkovic, zivkovic2006efficient}.

Figure 3 shows the effect of background subtraction on two frames: the 15000th frame of videos 13 and 14 (a-d and e-h respectively). In video 13 (a-d) the background does not change much over the course of the video and so all three background subtraction methods result in similar image foreground.  In video 14 a cloud moves onto the frame between the 3 min no-leak video used to generate the static background and the frame under analysis. In Figure 3f there is cloud in the image because fixed average background subtraction does not account for the recent movement of the cloud. However, Figure 3g and Figure 3h, implementing moving average and MOG-based background subtraction method respectively, treat the moving cloud as a part of the background and remove it in the foreground image. It is worth noting that all the background subtraction results are shown here in inverted colors for better visualization (the CNN does not use inverted colors).

\begin{figure}[!ht]
    \centering
    \includegraphics[width = 15cm]{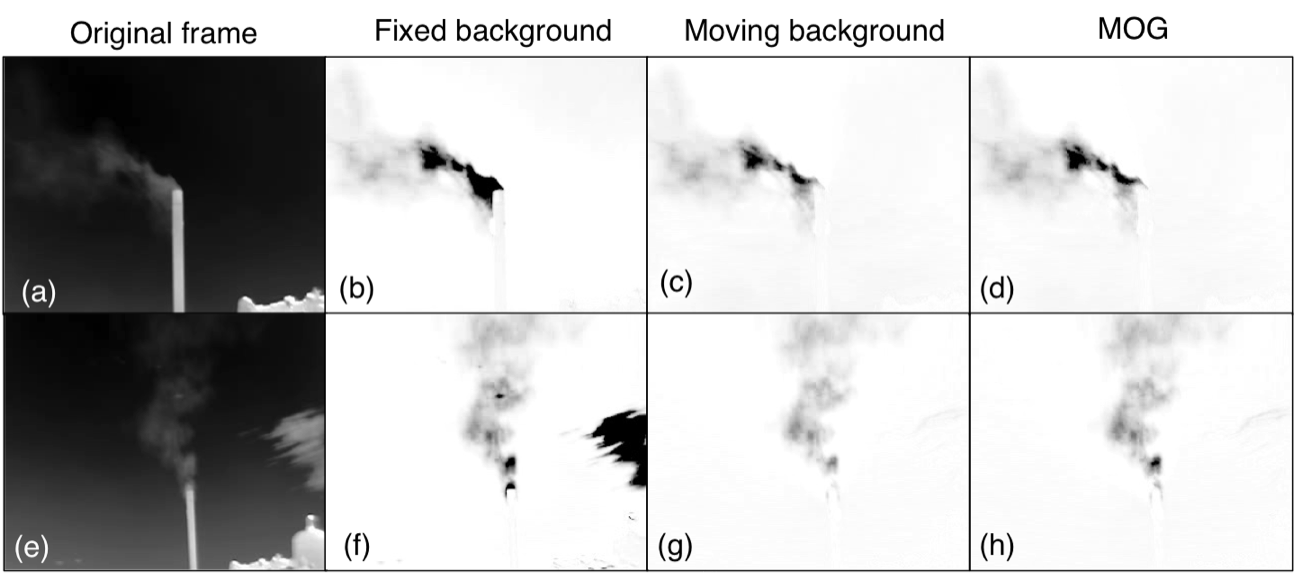}
    \caption{Top row - (a-d) shows the extracted plume after background subtraction using method (1)-(3) on the 15000th frame of video 13 (class-5 leak with leak rate: 84.2 scfh). The background does not change significantly across frames. Bottom row - (e-h) illustrates background subtraction results by method (1)-(3) on the 15000th frame of video 14 (class-5 leak with leak rate: 84.2 scfh) in which background shows substantial changes between frames. In video 14, the cloud does not exist in the non-leak segment, but the cloud starts moving in the leak segments. In Figure f there is moving cloud in the image because of performing fixed average background subtraction. But from Figure g and h, both method 2: moving average background subtraction method and method 3: MOG-based background subtraction method treat the moving cloud as a part of the background and remove it in the foreground image. It is worth noting that all the background subtraction results are in artificially inverted colors for better visualization.}
    \label{3}
\end{figure}

\subsection{Image normalization}
Image normalization is a crucial step in deep learning to ensure every image has similar pixel intensity distribution and to make the training algorithms converge faster compared to non-normalized input. For our image datasets, data normalization is performed by dividing every pixel in the image by 255, which is the maximum value of a pixel channel in the image.

\subsection{CNN model - GasNet}
In this paper, we develop our own deep CNN called GasNet, using TensorFlow software \cite{tersorflow}. GasNet is developed to perform binary image classification that distinguishes between non-leak images and leak images. Our CNN construction process follows general methods of building CNN models. Input images pass through some number of Conv-Pool structures and some number of fully-connected layers. Each Conv-Pool structure contains a convolutional layer, Rectified Linear Unit (ReLU) activation function, dropout regularization, a max pooling layer and batch normalization. In the convolutional layer, the input image is convolved with 4 filters with a size of 3x3. The ReLU activation function introduces nonlinearity to the network \cite{231n, nair2012}. Dropout is a regularization method where randomly selected neurons are updated or removed during training \cite{231n, hinton2012improving}. This can help prevent the CNN from overfitting \cite{231n, hinton2012improving}. 2x2 max pooling is implemented to reduce the spatial size of the image representation, the number of parameters in the network and the computational effort \cite{231n, ciresan2012}. Batch normalization is used after convolution, which makes the model train faster and increases its robustness \cite{231n, ioffe2015}. The first Conv-Pool structure contains 4 filters while the second contains 8 filters.

After these two Conv-Pool structures, the processed input image is passed through two fully-connected layers. The first one contains 2400 neurons; the second fully-connected layer contains 32 neurons and generates outputs for two classes (non-leak and leak). Then a softmax function is used to produce two probability scores for two classes and determine the prediction label (0 for non-leak, 1 for leak).

\begin{figure}[!ht]
    \centering
    \includegraphics{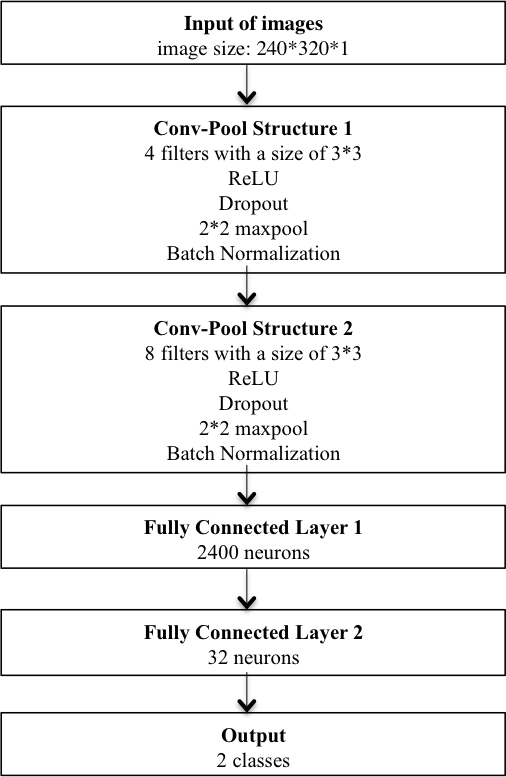}
    \caption{GasNet-2 deep CNN network architecture diagram}
    \label{4}
\end{figure}

We test three different CNN architectures. Figure 4 shows the structure of the moderate complexity version of GasNet, called GasNet-2. We also construct a simpler version called GasNet-1, which has one Conv-Pool structure and one fully-connected layer, and a more complex version called GasNet-3, which contains four Conv-Pool structures and two fully-connected layers.

\subsection{CNN model setup}
The Adam optimizer, which is an extension to stochastic gradient descent, is used to calculate the adaptive learning rate for each parameter \cite{ruder2016}. 

Table 2 illustrates how we split the training, validation and test dataset in each experiment. Training data are used for developing the model; validation data are used for tuning the hyperparameters and balancing the bias and the variance; test data are used to report final accuracy.  80\% of the data from the separator 2 are treated as training data, and the remaining 20\% of the data are treated as validation data. The test data are never introduced when we train the CNN model, and the test data are from a different piece of equipment (separator 1), taken at different times of day, camera orientation, etc. than the training and validation data.

\begin{table*}[!ht]
    \centering
    \caption{Training, validation and test size for each binary classification case and each distance}
    \label{2}
    \begin{tabular}{ >{\centering\arraybackslash}m{1in} | >{\centering\arraybackslash}m{1.5in} | >{\centering\arraybackslash}m{1.5in} | >{\centering\arraybackslash}m{1.5in}}
    \hline
     \textbf{Distance (m)}& \textbf{Size: training [frames]}& \textbf{Size: validation [frames]}& \textbf{Size: testing [frames]} \\
     \hline
     4.6 & 11369 & 2843 & 9466 \\
    6.9 & 11362 & 2841 & 9467\\
    9.8 & 11089 & 2773 & 9480\\
    12.6 & 11361 & 2841 & 9476\\
    15.6 & 11367 & 2842 & 9483\\
    \hline
    \end{tabular}
\end{table*}

\begin{table*}[h]
    \centering
    \caption{Illustration of detection accuracy calculation. Accuracy equals to the sum of true positive fraction and true negative fraction among all the test data.}
    \label{tab:my_label}
    \begin{tabular}{c|c|c}
    \hline
         & \textbf{Prediction: non-leak (0)} & \textbf{Prediction: leak (1)}\\
         \hline
        True class: non-leak (0) & True negative (TN) & False negative (FN) \\
        True class: leak (1) & False positive (FP) & True positive (TP) \\
        \hline
    \end{tabular}
\end{table*}

Accuracy is defined as the fraction of correct binary predictions of leaks and non-leaks (the sum of true positive fraction and true negative fraction among all the test data). We also compute the error bar of each accuracy. Because the test dataset is large, we split it randomly into 10 folds. Ten accuracy results are generated by performing the testing process on each of those test set folds using the same best trained model. The error bar shows the standard deviation of accuracy across the 10 test set folds.

\subsection{Baseline model method}
In order to have a model to compare to our CNN, we construct a method that does not use deep learning. We use optical flow analysis to calculate a baseline measure of methane plume detection efficacy \cite{kopilovic2000application, toreyin2005wavelet}. Optical flow estimates the apparent motion of objects between two consecutive frames \cite{opencvflow}. In particular, we use Gunner Farneback’s algorithm - a dense optical flow algorithm - to compute optical flow for all points in the image \cite{farneback}. The regions found to be moving will be assumed to be plume regions. The baseline method applies the moving average background subtraction method first and then uses the same setting of training data, validation data, and test data as those in deep-learning-based models. The baseline method is performed in four steps:
\\(1) Using the training set, we tune the parameters in Gunner Farneback’s algorithm in order to make the moving regions match the plume area visually.
\\(2) Then we determine two thresholds for this analysis. The first threshold is the movement magnitude threshold (MMT). If the estimated motion speed of a pixel is above the MMT, we consider the pixel as a moving pixel. Moving pixels are assumed to be plume pixels. The second threshold is the plume area threshold (PAT). If the number of moving/plume pixels is larger than the PAT, it indicates a leak plume in the image; if the plume area is smaller than the PAT, it indicates an image with no leak plumes. By looking at the motion speed distribution and the distribution of plume areas in the training data, we determine ranges of these two thresholds to be explored.
\\(3) Using the validation set, we loop over the ranges of MMT and PAT observed in the training set videos. We select during validation the threshold pair with the highest validation set prediction accuracy.
\\(4) We use the selected best threshold pair to make leak class predictions on test set and report the accuracy results.

\section{Results and discussion}
Below we show the results of our algorithms for the 7 leak-no-leak classification tasks, each performed across 5 imaging distances. In each case, we plot the accuracy of prediction on the y-axis, starting at 0.5. Because we test each algorithm on a set of 50\% leaks, 50\% non-leaks, a randomly guessing algorithm (``coin flipping'') assigning frames to leaking and non-leaking states would be expected to be correct 50\% of the time.

In the first set of results, we examine the effect of background subtraction on the accuracy of the moderate complexity CNN GasNet-2. In the second set of results we perform the sensitivity analysis on CNN architecture complexity. In the third set of results, we analyze three model training aggregation methods on different datasets.

\subsection{Results of different background subtraction methods}
First we explore the impact of background subtraction method, holding CNN complexity constant.

\begin{figure}[!ht]
    \centering
    \includegraphics[width = 15cm]{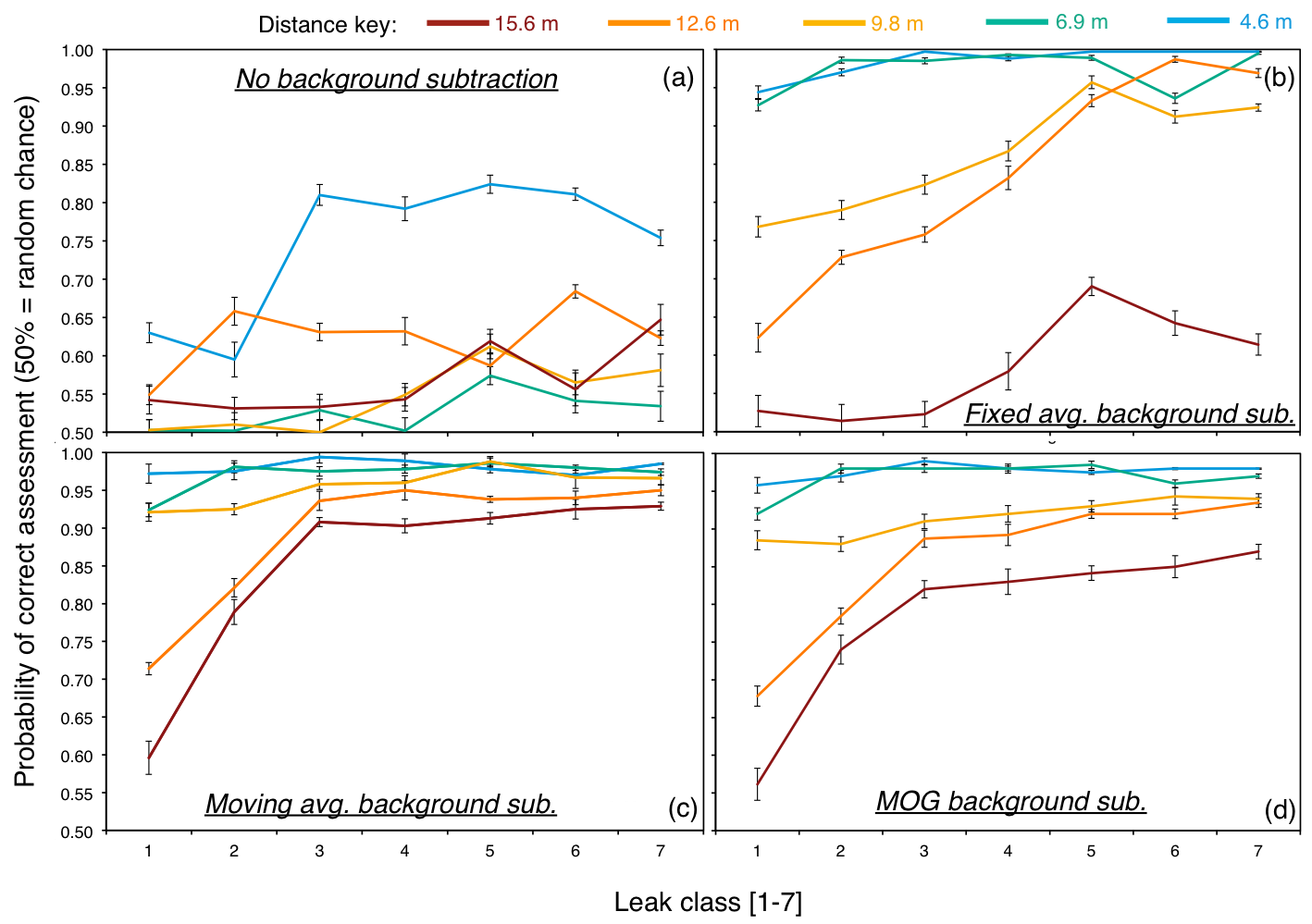}
    \caption{Probability of correct assessment by training 35 independent GasNet-2 CNN architectures on images with (a) no background subtraction method and three background subtraction methods: (b) fixed average background subtraction, (c) moving average background subtraction, and (d) MOG background subtraction. For all methods, accuracy improves with larger leak class. In all cases, a randomly guessing model would obtain an accuracy of 50\%. The error bars correspond to the standard deviation of accuracy across the 10 test set folds.}
    \label{5}
\end{figure}
Figure 5 shows the results of the GasNet-2 algorithm applied to the images with three varying background subtraction methods. For the without background subtraction method (Figure 5a), at the closest imaging distance (4.6 m), higher leak size generally leads to higher possibility of accurate assessment. The highest accuracy observed is 82\% (we round the accuracy result to the nearest percent because usually standard deviation is $\sim$1-2\%). The standard deviations are all below 2.0\%, suggesting that the trained CNN performs similarly across the 10-folds of the training data. The plume in the videos taken 4.6 m away from the leak source takes up most of the field of view of the camera, and the non-leak images contain mostly the background sky, so binary classification is feasible even without background subtraction. However, performance quickly degrades to near random chance with distance and decreasing leak size.

For the fixed average background subtraction (Figure 5b), all the accuracy levels exceed 93\% for imaging distances of 4.6 m and 6.9 m with small standard deviation ($\sim$1\%). At 9.8 m and 12.6 m, the corresponding curves show that it is much harder to differentiate between non-leak and class 1-3 leak sizes, which results in much lower accuracy compared with 4.6 m and 6.9 m. The accuracy increases with increasing leak size at the first four imaging distances.

The moving average background subtraction (Figure 5c) generates improved accuracy compared to the fixed average background subtraction method.  At an imaging distance of 4.6 m, the accuracies of all 7 leak combinations are above 97\%; similarly, at 6.9 m, the accuracies for differentiating non-leak with leak class of 2-7 are all above 98\%. In these two cases, the accuracy can reach as high as 99\%. There is a wider divergence between the curve representing 12.6 m, 15.6 m. Even at these distances, the accuracy of differentiating non-leak from leak exceeds 90\% for classes 3-7. The standard deviations across the 10 test set folds of all the 35 cases are lower than 1.6\%, except the leak class 1 at the farthest distance.

MOG-based background subtraction (Figure 5d) performs better than the fixed background subtraction method, but inferior to the moving average background subtraction method. There is little difference between MOG and moving average methods in the closest two measurements, namely 4.6 m and 6.9 m imaging distance. However, when the imaging distance exceeds 10 m, the accuracy gap between these two methods gets wider. On average, at the distance of 9.8 m, 12.6 m and 15.6 m, the accuracies of MOG-based background subtraction method are 3.9\%, 3.3\% and 6.4\% lower than those of moving average background subtraction method respectively.

Several conclusions can be drawn from Figure 5. First, it is critical to implement background subtraction before feeding data into the CNNs. A large amount of unrelated background information makes the CNN predictions perform poorly. Second, it is necessary to choose a proper background subtraction method. In our problem, where the videos are in grayscale and full of motions of non-rigid body (plume), moving average background subtraction works best compared to other methods. However, MOG-based background subtraction have been found effective in the literature \cite{friedman1997image, stauffer1999adaptive}. Given our limited dataset, MOG-based background subtraction method may perform better than moving average in some cases. By comparing fixed background subtraction and moving average background subtraction, the moving average background subtraction method generates better performance. This is likely due to the background changing over time in the video. The moving average background subtraction is a much more effective way to solve the problem of background change across video frames, which is particularly true at longer distances where more non-plume movements would be in the field of view for a fixed lens. 

Examining the cases where even the moving average background subtraction method provides relatively lower accuracy could help explain why the GasNet fails in some binary classification problems. For example, at the distance of 15.6 m, the accuracy for separating non-leak from class-1 leak is only 56\%. This is only slightly better than the random chance of 50\% correct. By human examination, we find it very difficult to observe these leaks, so the CNN likely has little signal to train on.

\subsection{Results of different architecture}
Using the moving average background subtraction method, we next analyze the sensitivity of our results to changing the deep learning architecture. We examine three different architectures - GasNet-1, GasNet-2 and GasNet-3, explained in section 4.4. We also use the baseline method of optical flow and plume area thresholding, again applied to the frames processed with moving average background subtraction.

When the imaging distance is small, the plume signal in the images is strong, making the detection problem easier. From Figure 6a, we can see that accuracy of the results from the three CNN architectures are all high, indicating that the difference among the three architectures is not significant. 

\begin{figure}[!ht]
    \centering
    \includegraphics[width = 15cm]{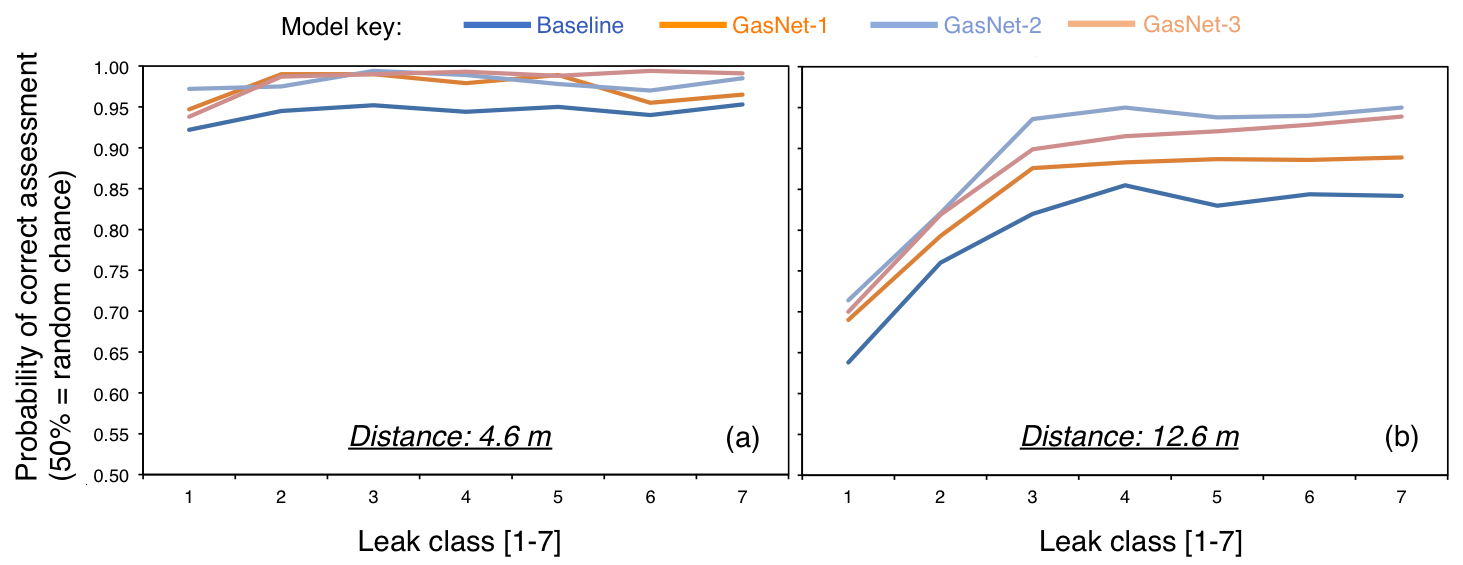}
    \caption{Model accuracy comparison after performing moving average background subtraction at the distance of 4.6 m (a) and 12.6 m (b) for the 3 CNN architectures and the baseline non-CNN architecture.}
    \label{6}
\end{figure}

When the imaging distance is large, the signal of whether there is a plume in the image is relatively weak, adding difficulties to the detection problem. From Figure 6b, it is clear that GasNet-2 outperforms the other two architectures. GasNet-1, which has one Conv-Pool structure and one fully-connected layer, is not complex enough to learn all the key features for the purpose of differentiating non-leak and leak. The number of parameters in GasNet-3, which has four Conv-Pool structures and two fully-connected layers, is 15\% larger than that of GasNet-2. This suggests that GasNet-3 has at least the potential for more complexity and capability to produce higher detection accuracy than GasNet-2. However, the performance of GasNet-3 does not improve much over GasNet-2 and even degrades at longer distance, 12.6 m. GasNet-3 may lack sufficient data to properly train its larger number of parameters.

In both cases, baseline results have the worst performance, indicating that CNN-based approach is sufficiently sophisticated to extract essential features from images to differentiate non-leak and leak and works better than a more traditional optical-flow-based analysis. In conclusion, GasNet-2, which is a moderate-depth, moderate-complexity CNN model variant, performs best across distances.

\subsection{Results of different model training aggregation methods}
We will use both the moving average background subtraction method and the GasNet-2 architecture in the following analysis, because the above combination generally works best based on the results of Section 5.1 and Section 5.2. We train our models using the following three methods:
\\Method 1: train 35 models, one for each distance and leak size combination.
\\Method 2: train 5 models, one for each distance with all leak sizes. 
\\Method 3: train 1 model for all combinations of leak size and imaging distance.

The performance results of models trained using method 1-3 are compared in Table 4. In addition to the accuracy at individual imaging distances, we also report the average accuracy across all distances. At each distance, a single model trained using method 2 is $\sim$1.5\% better than the average detection accuracy by 7 small models trained using method 1. By method 2, the detection accuracy is over 97\% at distances shorter than 9.8 m. We also achieve a detection accuracy of 95\% for all distances by method 3, higher than the 94\% average detection accuracy of the 5 models by method 2. We also apply the single model by method 3 to generate the detection accuracy for each distance. In general, method 2 is better than method 1 at each distance; method 3 is better than method 2 for all distances and leak sizes; method 3 can significantly increase the detection accuracy for distance 12.6 m and 15.6 m, which are poorly performed by method 1 and method 2. 

\begin{table*}[!ht]
    \centering
        \caption{Comparison of detection accuracy using three different model training aggregation methods. All the results are computed by GasNet-2 architecture after performing the moving average background subtraction method.}
    \begin{tabular}{c|c|c|c}
    \hline
      \textbf{Distance (m)}  & \textbf{Method 1} & \textbf{Method 2} & \textbf{Method 3} \\
      \hline
    4.6 & 98\% & 99\% & 98\% \\
    6.9 & 97\% & 99\% & 99\% \\
    9.8 & 89\% & 97\% & 96\% \\
    12.6 & 85\% & 91\% & 93\% \\
    15.6 & 85\% & 86\% & 91\% \\
    All distances & 93\% & 94\% & 95\%\\
    \hline
    \end{tabular}
\end{table*}

\begin{figure}[!ht]
    \centering
    \includegraphics[width = 15cm]{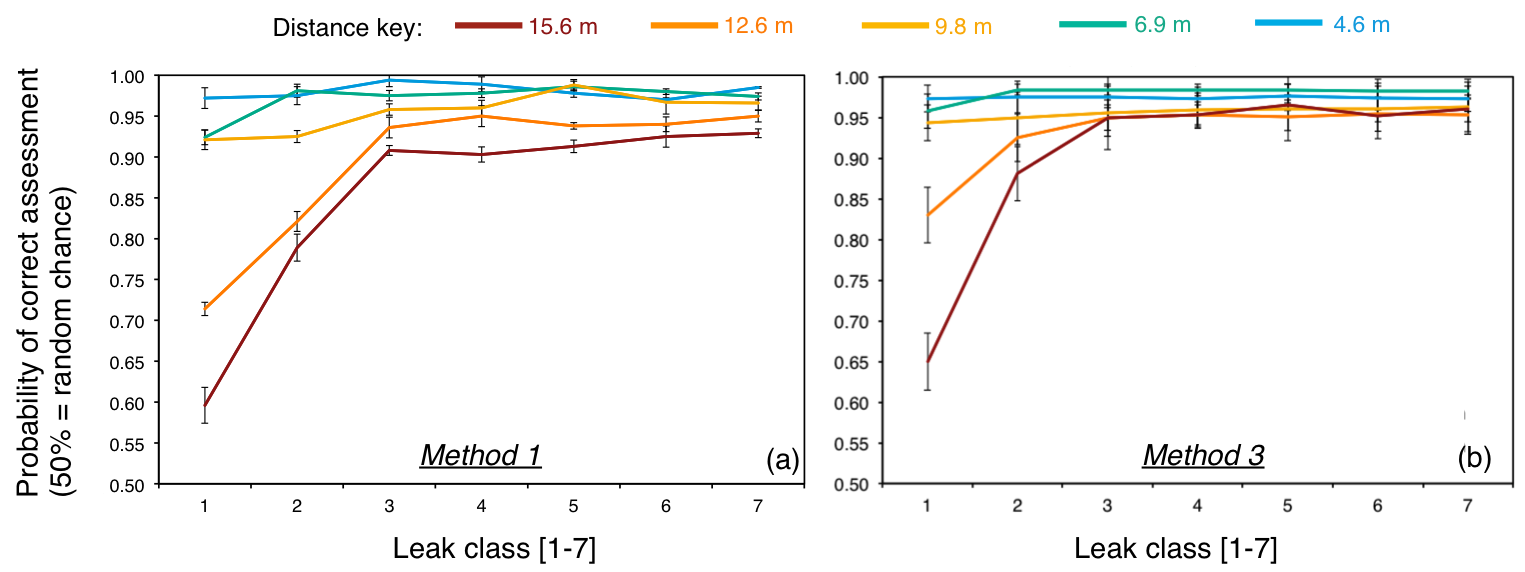}
    \caption{The probability of correct assessment using method 1 and method 3. Method 1 is training 35 models, one of which is for each distance and leak size combination; method 3 is training 1 model for all combinations of leak size and imaging distance. All the results are computed by GasNet-2 architecture after performing the moving average background subtraction method. Figure 7a is the same as Figure 5c.}
\end{figure}

Now with the use of the moving average background subtraction method, the GasNet-2 architecture, we generate the curves of probability of correct assessment by method 3, which is shown in Figure 7b. As a comparison to Figure 7a by method 1, method 3 moves up the curves representing 12.6 m and 15.6 m by 3.8\% and 6.1\% respectively. From Figure 7b, binary detection accuracy is over 94\% across all leak sizes at close distances ($\sim$5-10 m). At farther distances ($\sim$13-16 m), leaks larger than class 3 are detected with an accuracy over 95\%. 

From Figure 7 and Table 4, we can see that method 3 produces the best results. This may be because increasing the training data size and seeing data with more variety in method 3 may help reduce model variance generalization error. 

In the end, we find that the imaging distance is the most important parameter affecting the effectiveness of automated OGI-based technology. This is consistent with what Ravikumar et, al. found by simulations and experiments in \cite{ravikumar2016optical, ravikumar2018good}. From our results, the curve representing the assessment accuracy across 7 combinations of non-leak and leak shows that generally the curve shifts lower as the imaging distance increases. There are fewer plume pixels to use when the imaging distance is greater.  In particular, we can see a dramatic decrease in detection accuracy at the smallest leak sizes and longest distances.

\section{Concluding remarks}
Our paper illustrates a novel application of deep learning and computer vision that have a great potential to tackle the important environmental problem of methane emissions and simplify the leak survey and detection with high accuracy. Without collecting a background image, background subtraction method combined with CNN-based GasNet algorithm is shown to be an appropriate approach for leak detection, which eventually can be trained to handle complex and real-world environments.

It is worth noting that the detection accuracy results are not comparable between the automated OGI-based detection technology presented in this study and OGI technology operated by people in prior work \cite{ravikumar2018good}. We cannot draw a conclusion that the algorithm's ability to identify leaks is not as good as a human. The main reasons are: in \cite{ravikumar2018good} operator used the high sensitivity mode of the IR camera, which is more sensitive to even weak movement of small leaks compared to the normal mode used in this study; and the presence of a leak in \cite{ravikumar2018good} was determined by observing a leak video where we analyzed individual frames in this study.

Our machine vision algorithm, which is the first one in the field of methane emission detection, is demonstrated to be successful using the GasVid dataset. Future work will be done to develop the technology and enlarge the GasVid dataset so that it can represent the diversity of leaks that are observed in the real world and make the GasNet algorithm more generalizable to real-world leaks. Currently, although the GasVid dataset was collected from only a test environment, the algorithm is expected to perform well in real-world leaks as well, because the background subtraction method allows the GasNet to focus on the leaks regardless of the leak location or background condition.

The accuracy of our algorithm will depend on the performance of the camera affected by imaging distance, temperature contrast, and movement of background, etc\cite{ravikumar2016optical}. We plan to examine these issues in future studies. 

Future work will also include exploration of different model architectures. For example, temporal information in the videos can also be analyzed and used to detect methane leaks using sequence models in order to capture plume motion. Given the number of environmental parameters that affect plume dispersion, hybridization of machine learning and physical models could lead to better automatic detection. Such physical models may help the model understand the flow characteristics of the plumes, and could help quantify plume volume flux rate from imagery.

In the future, the automated OGI-based technology could be equipped on roof of maintenance vehicles or in a “security camera” approach at perimeter of sites to achieve so-called “automatic vigilance”, saving the labor cost of manual IR detection and accelerating the process of leak detection and repair.

\section*{Acknowledgment}
This work was supported by the Stanford Interdisciplinary Graduate Fellowship and the Stanford Natural Gas Initiative.

\section*{Declaration of interest}
Declaration of interest: none.

\section*{References}

%\bibliography{mybibfile}

\end{document}